\documentclass[sigconf,nonacm]{acmart}
\usepackage{cleveref}
\usepackage{preprint_header}

\AtBeginDocument{%
  \providecommand\BibTeX{{%
    \normalfont B\kern-0.5em{\scshape i\kern-0.25em b}\kern-0.8em\TeX}}}

\begin{document}

\title{What's in the News? Towards Identification of Bias by Commission, Omission, and Source Selection (COSS)}


\author{Anastasia Zhukova$^{1,2}$, Terry Ruas$^{2}$, Felix Hamborg$^{3}$, Karsten Donnay$^4$, Bela Gipp$^{2}$}
\def \authors{author one, author two, author three, author four}
\affiliation{%
\institution{$^1$University of Wuppertal
\country{Germany}}
\streetaddress{zhukova@uni-wuppertal.de}
}
\affiliation{%
  \institution{$^2$University of G{\"o}ttingen
  \country{Germany}}
}
\affiliation{%
\institution{$^3$Heidelberg Academy of Sciences and Humanities
\country{Germany}}
}
\affiliation{%
\institution{$^4$University of Zurich
\country{Switzerland}}
}
\email{zhukova@gipplab.org}

\renewcommand{\shortauthors}{Zhukova et al.}

\begin{abstract}
In a world overwhelmed with news, determining which information comes from reliable sources or how neutral is the reported information in the news articles poses a challenge to news readers. In this paper, we propose a methodology for automatically identifying bias by commission, omission, and source selection (COSS) as a joint three-fold objective, as opposed to the previous work separately addressing these types of bias. In a pipeline concept, we describe the goals and tasks of its steps toward bias identification and provide an example of a visualization that leverages the extracted features and patterns of text reuse.
\end{abstract}






\maketitle
\thispagestyle{preprintbox}

\section{Introduction and Related work}
The literature on media bias has found that editorial choices in the news production process, such as bias by commission and omission of information and source selection (COSS), strongly affect public perceptions \cite{bakshy2015exposure}. This finding is particularly alarming since today's news production system faces pressure to minimize reporting costs \cite{fengler2008journalists, papacharissi2008news}. Consequently, journalists often rely on the same news source, copy reports, or the (factual) information in other reports \cite{frank2003these}. This phenomenon, also termed pack journalism, tends to lead to a lower quality of reporting, as journalists fail to independently verify the information they report \cite{matusitz2012examination}. 

Unlike scientific publications, where sources of information must be documented explicitly, news articles typically contain no citations \cite{christian2014associated}. However, much of the information in articles typically originates from previously published articles, newswire reports, or press releases \cite{gaizauskas2001meter, papacharissi2008news}. Compared to its sources, which information is included or excluded in an article is typically opaque to the news reader \cite{hamborg2019automated}. Especially when information is reused as paraphrases where different to the original source wording is used, which eventually leads to biased reporting \cite{hamborg2019automated2}. 

\begin{figure}
\centering
  \includegraphics[width=0.47\textwidth]{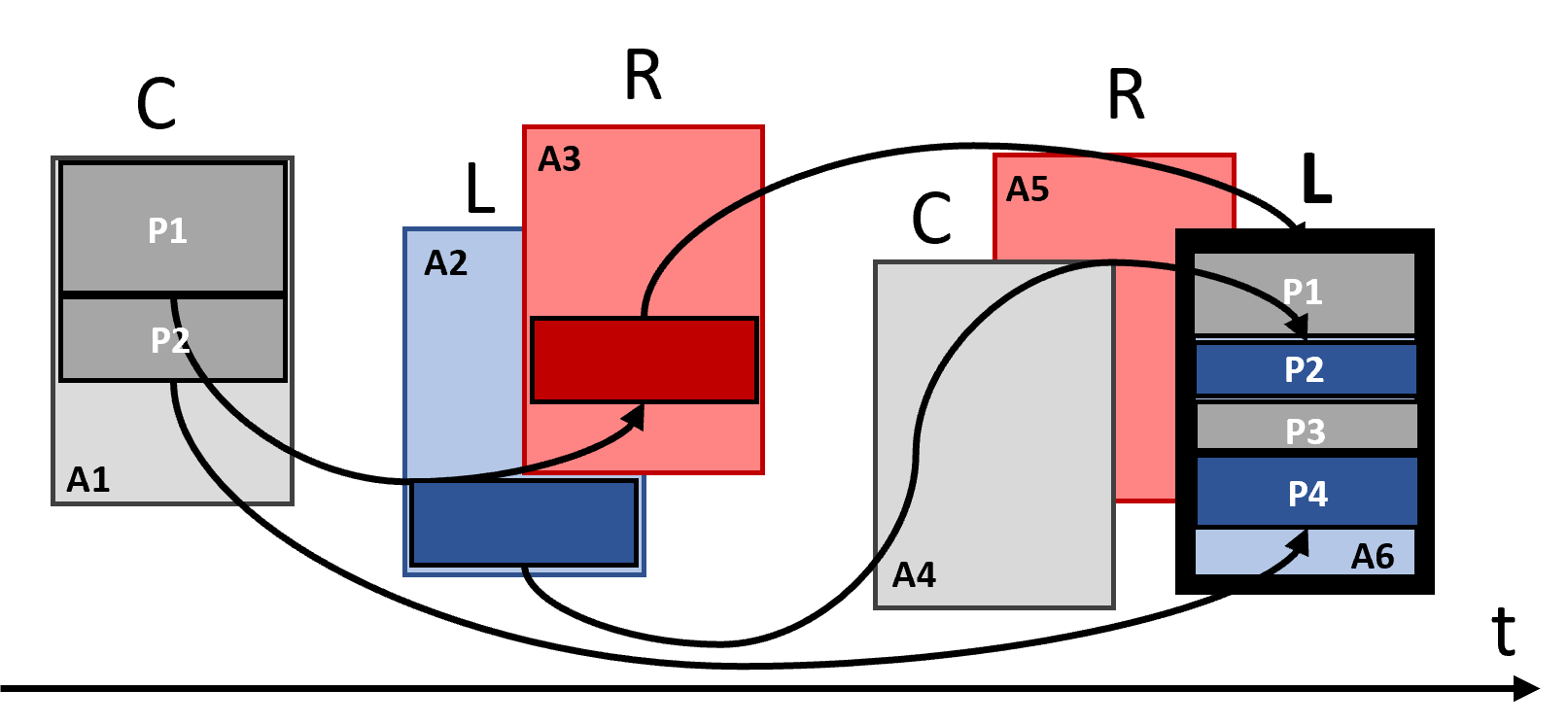}
  \caption{An example of visualizing the extracted information by pipeline for identifying bias by COSS. The identified text reuse in a set of related articles is ordered by date-time. The figure shows that a seed article A6 with left polarity (L) contains both original and paragraphs with reused information. For example, paragraph P3 contains original information labeled as center-oriented (C), although the article belongs to a left-oriented publisher. On the contrary, paragraph P4 was reused from article A1 and has changed its polarity from the original central to the left. }
  \label{fig:flow}
\end{figure}

In the past two decades, computer science has addressed the problem of automated identification of text reuse. Studies of text reuse substantially focus on (1) plagiarism detection, semantic textual similarity, and paraphrase identification, (2) text reuse in journalism, blog posts, and newswire reports, on the Web, and in Wikipedia, (3) information flow analysis, story diffusion and propagation, and news story chains, (4) novelty detection. However, the adaption and application of the automated analysis approach to the news domain and the study of bias are only just emerging \cite{hamborg2019automated}.

Previous approaches considered identifying bias by commission and omission as two separate tasks \cite{Ehrhardt2021,hamborg2021newsdeps}. They used statistical approaches that relied on direct text reuse, e.g., TFIDF, and focused only on one type of bias. Moreover, the existing independent news aggregators that cover a full political spectrum, such as AllSides\footnote{\url{https://www.allsides.com/}}, Ground.News\footnote{\url{https://ground.news/}}, and The True Story\footnote{\url{https://thetruestory.news/}} either focus on one of these types of biases or perform the analysis on an article level. In this paper, we propose a methodology and a concept pipeline that identifies bias by COSS as a three-fold objective. We address text reuse on a more conceptual level, such as paraphrasing, and aim at identifying text reuse on a higher level of granularity, e.g., paragraphs.



\section{Pipeline Concept}
The pipeline for identification of bias by COSS has two purposes: (1) analysis of a given seed article against a collection of event-related articles to identify which parts of it are original and which are reused, (2) identification of patterns in information flows in a collection of event-related articles to explore a bigger picture on information reuse. Both of the flows require the same pipeline stages: (a) candidate retrieval, (b) source retrieval and text alignment, (c) construction of a graph of text reuse, (d) pattern analysis, (e) visualization of the extracted information. 

\textbf{Candidate retrieval} obtains articles reporting the same event. The step extracts event-related documents from a large database of news articles, e.g., LexisNexis\footnote{\url{https://www.lexisnexis.com/}}, CommonCrawl\footnote{\url{https://commoncrawl.org/}}, MediaCloud\footnote{\url{https://mediacloud.org/}}, and The GDELT Project\footnote{\url{https://www.gdeltproject.org/}}. To retrieve related documents, the system should use either an event-descriptive query and a time frame for this event or a seed document with its timestamp. Alternatively, for the system evaluation in a closed environment, candidate retrieval supports reading a provided set of related articles that contain all required attributes, e.g., a timestamp. Similar to Ground.News, to each article, we assign a polarity label induced from an outlet, e.g., each article from Fox News will be labeled as ``R'' for right or conservative slant.   

\textbf{Source retrieval and text alignment} are the core steps in identifying information reuse that analyze which parts of text are reused from which source(s). Unlike most existing methods for text alignment that identify copy-pastes or word permutations, we focus on identifying paraphrased sentences or paragraphs that convey the same message but use different and possibly loaded wording. 

\textbf{Polarity classification} enables revisiting both original and reused paragraphs and checks if the outlet-inferred labels correspond to the labels from a polarity classifier. Such a polarity relabeling tracks how the same message evolves over time and across the outlets (see \Cref{fig:flow}). Training a reliable classifier requires a large balanced dataset to incorporate variance of the biased language \cite{gebhard2020polusa}.

\textbf{A graph of text reuse} stores articles, their paragraphs, and both extracted and assigned attributes to enable exploration of the patterns of information reuse. The relations between the paragraphs encode the strength of semantic similarity between the paragraphs, and the time codes of the articles enforce a directed graph, which is required for source identification. 

\textbf{Statistical and network analysis} aims at identifying \textbf{patterns} of text reuse that may induce bias by COSS. Analysis of the article's origin includes determining how many paragraphs originate from which sources with which polarities. For example, if an article reuses a significant amount of information from neutral sources, such as news agencies, we could conclude that this article is reliable and unlikely bias-prone. On the contrary, if an article consists of too many slanted paragraphs with no sources, it might indicate a lack of trustworthiness in this article. Analyzing the information for bias by the commission includes analysis of which information with which polarity tends to be reused, how often and how reused paragraphs change polarity, how long information continues being reused after the original publishing, etc. On the contrary, analyzing patterns of bias by omission includes identifying which parts of the source articles were not picked up by which articles were excluded from discussions. For example, if a source article is left-oriented, a right-oriented article could reuse only excerpts that report about an event itself but omit parts that fall into the liberal agenda. 

\textbf{Visualization} is an efficient way to enable researchers and news readers to explore the extracted information in a tempo-oriented graph structure (see \Cref{fig:flow}). Additionally, visualization depicts such identified features as the strength of semantic similarities between paragraphs \cite{hamborg2021newsdeps}, the original and assigned polarity of paragraphs and articles, and highlighted patterns that may indicate biases of each of the three types.

\section{Conclusion}
In this paper, we propose a concept of identification of bias by COSS as a joint three-fold objective. Compared to the previous systems identifying these types of bias via simple direct text reuse, our system leverages a solid research basis in text plagiarism detection and recent advances in paraphrase identification with semantic similarity. Revealed cases of paraphrasing help determine the source of articles and how reused information changed over time and news articles.





\bibliographystyle{ACM-Reference-Format}
\bibliography{sample-base}










\end{document}